\documentclass[10pt,twocolumn,letterpaper]{article}

\usepackage{iccv}
\usepackage{times}
\usepackage{epsfig}
\usepackage{graphicx}
\usepackage{amsmath}
\usepackage{amssymb}

\usepackage{multirow}
\usepackage{makecell}
\usepackage{subfigure} 
\usepackage{bbding}
\usepackage{algorithm}
\usepackage{algorithmic}
\usepackage{stmaryrd}
\usepackage{booktabs}
\usepackage[pagebackref=true,breaklinks=true,letterpaper=true,colorlinks,bookmarks=false]{hyperref}

\iccvfinalcopy % *** Uncomment this line for the final submission

\def\iccvPaperID{****} % *** Enter the ICCV Paper ID here

\ificcvfinal\fi

\begin{document}

%%%%%%%%% TITLE
\title{One-Nearest Neighborhood Guides Inlier Estimation for Unsupervised Point Cloud Registration}

\author{
	Yongzhe Yuan\textsuperscript{\rm 1}
	\quad
	Yue Wu\textsuperscript{\rm 1}\thanks{Corresponding authors.}
	\quad
	Maoguo Gong\textsuperscript{\rm 2}
	\quad
	Qiguang Miao\textsuperscript{\rm 1}
	\quad
	A. K. Qin\textsuperscript{\rm 3}\\
\textsuperscript{\rm 1}School of Computer Science and Technology, Xidian 
University, China\\
\textsuperscript{\rm 2}School of Electronic Engineering, Xidian University, 
China\\
\textsuperscript{\rm 3}Swinburne University of Technology, Australia\\
{\tt\small yyz@stu.xidian.edu.cn, \{ywu, qgmiao\}@xidian.edu.cn,  gong@ieee.org, kqin@swin.edu.au}
}

\maketitle
\ificcvfinal\thispagestyle{empty}\fi

%%%%%%%%% ABSTRACT
\begin{abstract}
   The precision of unsupervised point cloud registration methods is typically limited by the lack of reliable inlier estimation and self-supervised signal, especially in partially overlapping scenarios. In this paper, we propose an effective inlier estimation method for unsupervised point cloud registration by capturing geometric structure consistency between the source point cloud and its corresponding reference point cloud copy. 
   Specifically, to obtain a high quality reference point cloud copy, an One-Nearest Neighborhood (1-NN) point cloud is generated by input point cloud. This facilitates matching map construction and allows for integrating dual neighborhood matching scores of 1-NN point cloud and input point cloud to improve matching confidence. 
   Benefiting from the high quality reference copy, we argue that the neighborhood graph formed by inlier and its neighborhood should have consistency between source point cloud and its corresponding reference copy. Based on this observation, we construct transformation-invariant geometric structure representations and capture geometric structure consistency to score the inlier confidence for estimated correspondences between source point cloud and its reference copy. 
   This strategy can simultaneously provide the reliable self-supervised signal for model optimization. Finally, we further calculate transformation estimation by the weighted SVD algorithm with the estimated correspondences and corresponding inlier confidence. We train the proposed model in an unsupervised manner, and extensive experiments on synthetic and real-world datasets illustrate the effectiveness of the proposed method. 
\end{abstract}

%%%%%%%%% BODY TEXT
\section{Introduction}

With the rapid development of 3D data acquisition technology, point cloud data collected by LiDAR 
\cite{zhang2014loam}, Structured Light Sensors \cite{salvi2004pattern}, and Stereo Cameras \cite{engel2015large} has become ubiquitous in various 3D computer vision and robotics applications \cite{pomerleau2015review, wang2019densefusion}, such as 
autopilot \cite{geigerwe}, surgical navigation \cite{ma2022augmented}, and simultaneous localization and mapping \cite{fioraio2011realtime}. In such applications, rigid body point cloud registration plays an essential role, which aims to find a rigid transformation to align one point cloud to another.

\begin{figure}
	\centering
	\subfigure[]{
		\includegraphics[width=3.3in]{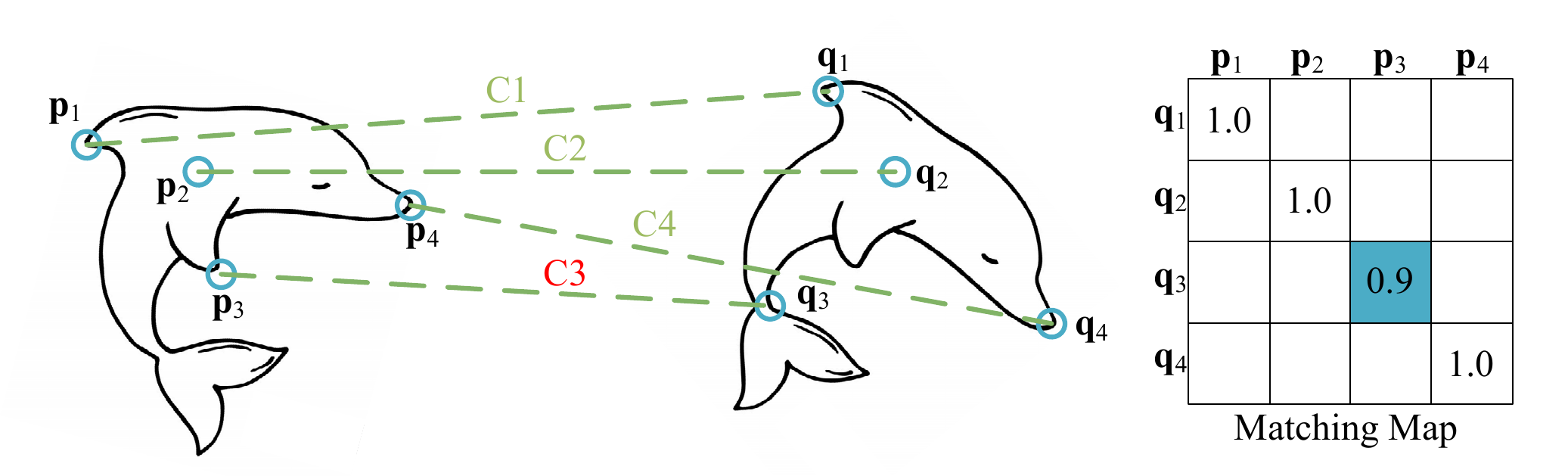}
		\label{introda}
	}
	\subfigure[]{
		\includegraphics[width=3.3in]{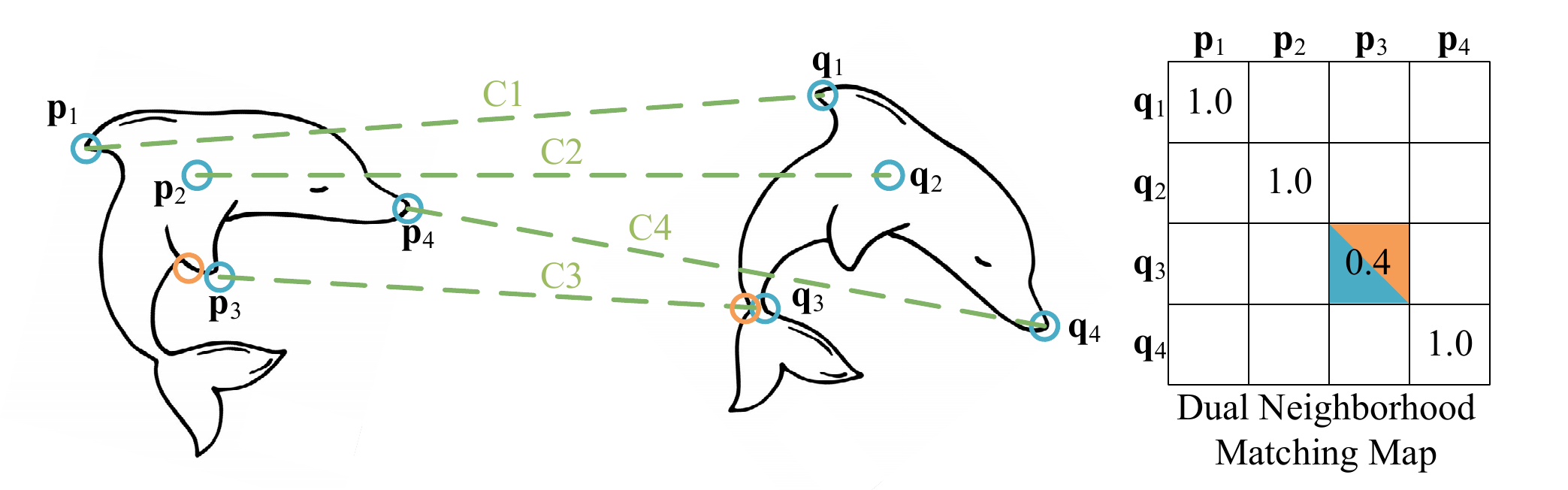}
		\label{introdb}
	}
	\caption{A toy example of \textcolor[RGB]{75,172,198}{single neighborhood} and proposed \textcolor[RGB]{245,157,86}{1-NN strategy}. (a) shows that false matching (C3) has very high matching score because of high \textcolor[RGB]{75,172,198}{single neighborhood} similarity and similar contextual feature. (b) shows that the matching score is adjusted with the help of \textcolor[RGB]{245,157,86}{1-NN strategy}, and C3 is correctly judged as a false matching.} 
	\label{introduction-dual}
\end{figure}

The recent advances have been dominated by learning-based methods. Most of these methods focus on solving the point cloud registration task in a supervised manner \cite{li2020iterative, yuan2020deepgmr, wang2019deep, huang2020feature, yew2020rpm}. They require labeled data as the supervision signal to learn effective representations. However, obtaining labeled data is cumbersome and time-consuming, which may hinder applications in real scenarios. To address this limitation, unsupervised point cloud registration methods gradually attract scholars' attention. Global alignment difference is a widely used optimization signal in unsupervised methods, which learns the optimal rigid transformation to align point cloud pair perfectly by minimizing Chamfer Distance \cite{marcon2021unsupervised, zhao2021mm}. Nevertheless, Chamfer Distance is very sensitive to the presence of outlier and cannot provide effective self-supervised signal to guide inlier estimation on partially overlapping point cloud.

To tackle issues mentioned above, we propose an effective inlier estimation method for unsupervised point cloud registration by capturing geometric structure consistency between source point cloud and its corresponding reference point cloud copy. 
Two crucial questions remain to be addressed in order to make unsupervised inlier estimation a success: \textit{How to obtain a high quality corresponding reference point cloud copy?} \textit{Can single neighborhood strategy provide reliable matching map to generate reference copy?}

We provide answers to both questions. Our key insight is that relying solely on single neighborhood is unreliable to generate matching map.
As shown in Figure \ref{introda}, the false matching C3 is mistakenly identified as a correct one due to high single neighborhood similarity and similar contextual feature. Interestingly,  
introducing the closest point neighborhood of raw point to aid judgement can greatly alleviate this dilemma.
The reason is that at least one of raw point and closest point has correct matching. Even if two points are both false matching, integrated dual neighborhood matching score in the matching map is typically quite small and the real impact is limited. We define this strategy as One-Nearest Neighborhood (1-NN) shown in Figure \ref{introdb}. The dual neighborhood matching score of C3 has decreased due to low matching score of 1-NN strategy and thus C3 is correctly judged as a false matching.

Motivated by the discussion above, we propose the dual neighborhood fusion matching module to facilitate matching map construction and integrate dual neighborhood matching scores to generate high quality reference point cloud copy. 
This module employs a 1-NN strategy, which selects the closest point from the input point cloud to generate a 1-NN point cloud.
Benefiting from the high quality reference copy, the neighborhood graph formed by inlier and its neighborhood should have consistency between source point cloud and its corresponding reference copy, and outlier is just the opposite. Based on this observation, we propose the geometric neighborhood inlier estimation module to construct effective transformation-invariant geometric structure representations and capture their consistency to
score the inlier confidence for each estimated correspondences between source point cloud and its corresponding reference copy. This module
provides simultaneously effective byproduct as self-supervised signal based on geometric structure for model optimization. 
To demonstrate efficacy of the proposed method, we conduct extensive experiments on the synthetic datasets ModelNet40 \cite{wu20153d}, Augmented ICL-NUIM \cite{choi2015robust} and real-world dataset 7Scenes \cite{zeng20173dmatch}. 
Experimental results illustrate 
capturing geometric structure consistency between the source point cloud and its corresponding reference copy is effective for inlier estimation.
To summarize, our contributions are as follows:
\begin{itemize}
	\item We propose the dual neighborhood fusion matching module to facilitate matching map construction, which can generate high quality reference copy for source point cloud.
	\item  Based on high quality reference copy, we design a geometric neighborhood inlier estimation module to score the inlier confidence for each estimated correspondences between source point cloud and its reference copy.
	\item In the unsupervised setting, instead of using the ground-truth transformation, we construct geometric structure consistency objective based on transformation-invariant self-supervised signal for model training and optimization.
\end{itemize}

%------------------------------------------------------------------------
\section{Related Work}
\textbf{Traditional point cloud registration methods.} 
Most traditional methods need a good initial transformation and
converge to the local minima near the initialization point. One of the most profound methods is the Iterative Closest Point (ICP) algorithm \cite{besl1992method}, which begins with an initial transformation and iteratively alternates between solving two trivial subproblems: finding the closest points as correspondence under current transformation, and computing optimal transformation by SVD \cite{kurobe2020corsnet} based on identified correspondences. Though ICP can complete a high-precision registration, it is susceptible to the initial perturbation. In recent years, variants of ICP have been proposed \cite{segal2009generalized, yang2015go, bouaziz2013sparse, fitzgibbon2003robust, rusinkiewicz2019symmetric}, and they can improve the defects of ICP and enhance the registration accuracy \cite{bellekens2015benchmark}. However, these methods retain a few essential drawbacks. Firstly, they depend strongly on
the initialization. Secondly, it is difficult to integrate them into the deep learning pipeline as they lack differentiability. Thirdly, explicit estimation of corresponding points leads to quadratic complexity scaling with the number of points \cite{rusinkiewicz2001efficient}, which can introduce significant computational challenges.

\textbf{Learning-based registration methods.} 
At present, most learning-based methods are based on supervision \cite{wang2019deep, yuan2020deepgmr, qin2022geometric, fischer2021stickypillars}. PointnetLK \cite{aoki2019pointnetlk} is a classical correspondence-free method, which calculates global feature descriptors through PointNet and iteratively uses the Inverse Compositional formulation and LK algorithm (IC-LK) \cite{lucey2012fourier, lucas1981iterative} to minimize distance between the descriptors to achieve registration. RPM-Net \cite{yew2020rpm} utilizes the differentiable Sinkhorn layer and annealing to get soft assignments of point correspondences from hybrid features learned from both spatial coordinates and local geometry.

Recently, unsupervised point cloud registration has gained increasing attention due to its applicability in scenarios where labeled training data is scarce or unavailable. Some methods have been proposed to address this challenge and achieved promising results \cite{kadam2020unsupervised, feng2021recurrent, el2021unsupervisedr, bauer2021reagent, jiang2021sampling}. Feature-metric point cloud registration framework (FMR) \cite{huang2020feature} enforces the optimisation of registration by minimising a feature-metric projection error with a autoencoder-based network. Unfortunately, the registration performance will significantly decline on the partial data due to the lack of inlier estimation. RIE \cite{shen2022reliable} propose a 
inlier estimation method, which can capture graph-structure difference between source point cloud and the reference point copy generated by single neighborhood strategy for inlier estimation. However, single neighborhood strategy restrains high quality reference point copy generating possibly and thus affect inlier estiamtion.
In comparison, our method achieves stable and reliable reference point copy generating with 1-NN for inlier estimation and can provide effective self-supervised signal for model optimization.

%------------------------------------------------------------------------
\section{Methodology}
\begin{figure*}
	\centering
	\includegraphics[width=7.2in]{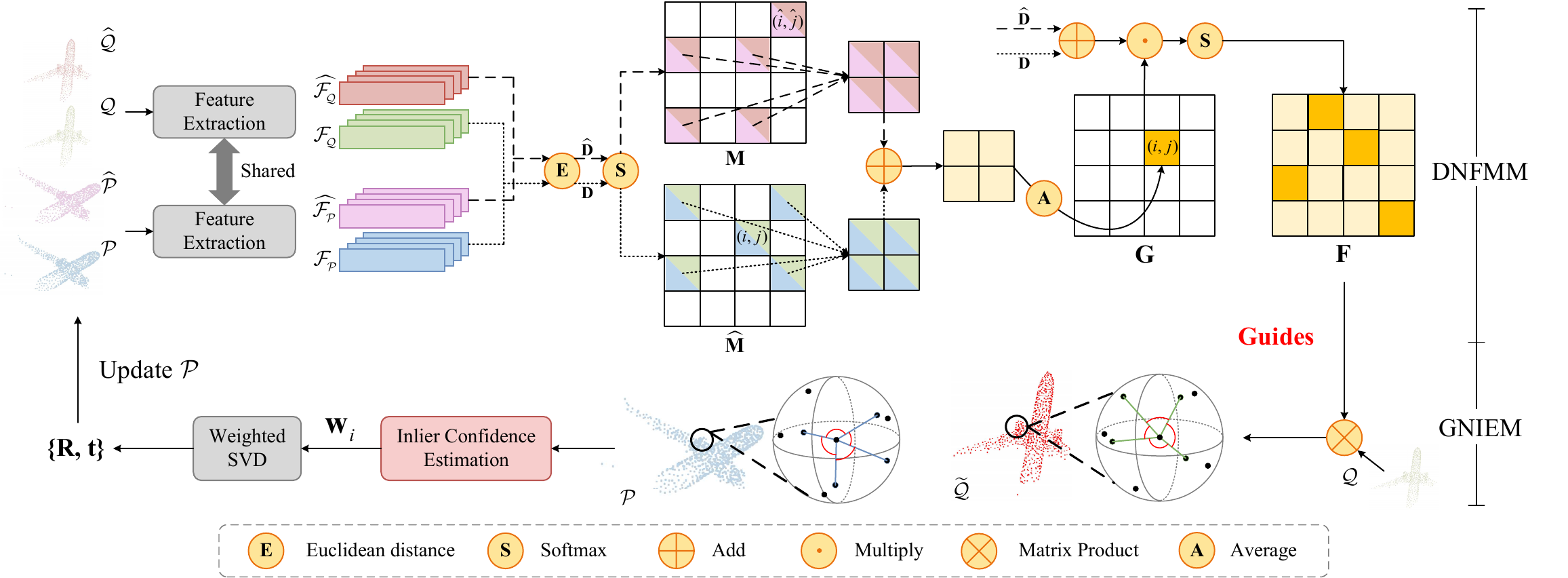}
	\caption{The pipeline of the proposed method. We first extract local features of input point cloud and 1-NN point cloud with DGCNN. The dual neighborhood fusion matching module (DNFMM) aims to facilitate the construction of matching map $\textbf{F}$ and improve matching confidence for generating high quality reference point cloud copy $\widetilde{\mathcal{Q}}$. Based on the observation that the neighborhood graph formed by inlier and its neighborhood should have geometric structure consistency between $\mathcal{P}$ and $\widetilde{\mathcal{Q}}$, the geometric neighborhood inlier estimation module (GNIEM) constructs transformation-invariant geometric structure representations (edge and included angle) and captures their consistency to score the inlier confidence for each estimated correspondences between $\mathcal{P}$ and $\widetilde{\mathcal{Q}}$.
	Finally, the transformation is estimated with SVD method via the reliable inliers correspondence.}
	\label{framework}
\end{figure*}

We first introduce notations utilized throughout this paper. Given two point clouds:
source point cloud $\mathcal{P}=\lbrace{\textbf{p}_i}\in\mathbb{R}^3\mid{i=1,\dots,N}
\rbrace$ and reference point cloud $\mathcal{Q}=\lbrace{\textbf{q}_j}\in\mathbb{R}^3\mid{j=1,\dots,M}\rbrace$,
where each point is represented as a vector of $(x, y, z)$ coordinates. 
Point cloud registration task aims to estimate a rigid transformation 
$\lbrace\textbf{R},\textbf{t}\rbrace$ which accurately aligns 
$\mathcal{P}$ and $\mathcal{Q}$, with a 3D rotation 
$\textbf{R}\in SO(3)$ and a 3D translation $\textbf{t}\in \mathbb{R}^3$.
The transformation can be solved by:
\begin{equation}
	\min_{\textbf{R},\textbf{t}}\sum_{(\textbf{p}_i^{*},\textbf{q}_j^{*})\in
		\mathcal{H}^{*}}\|{\textbf{R}\cdot\textbf{p}_i^{*}+
		\textbf{t}-\textbf{q}_j^{*}}\|_2^2,
	\label{eq:corr}
\end{equation}
where $\mathcal{H}^{*}$ is the set of ground-truth correspondences between
$\mathcal{P}$ and $\mathcal{Q}$. 
%In our work, we first extract point correspondences between source point cloud and pseudo reference point cloud and then estimate the alignment transformation by correspondences.
We propose an effective inlier estimation method for unsupervised point cloud registration by capturing geometric structure consistency between source point cloud and its corresponding reference point cloud copy. To obtain a high quality reference copy, we design dual neighborhood fusion matching module to facilitate matching map construction with 1-NN strategy. Benefiting from the high quality reference copy, we design the geometric neighborhood inlier estimation module to construct effective transformation-invariant geometric structure representations and capture their consistency to score the inlier confidence for each estimated correspondences between source point cloud and its reference copy. The pipeline is illustrated in Figure \ref{framework}.

\subsection{Dual Neighborhood Fusion Matching Module}
\label{nfmm}
The matching score is typically calculated by single point feature distance \cite{yew2020rpm} or single neighborhood integration \cite{shen2022reliable} when constructing matching map. These methods may suffer from false matching due to similar contextual feature, as illustrated in Figure \ref{introduction-dual}. In this paper, we propose the 1-NN strategy to generate 1-NN point cloud, which can alleviate this dilemma and facilitate matching map construction for obtaining high quality reference point cloud copy. 

We first define 1-NN point cloud $\widehat{\mathcal{P}}$ and $\widehat{\mathcal{Q}}$ as the closest point for each point in $\mathcal{P}$ and $\mathcal{Q}$, respectively:
\begin{equation}
	\begin{aligned}
		\widehat{\mathcal{P}}=\lbrace{\widehat{\textbf{p}_i}}\mid{{\widehat{\textbf{p}_i}}=\arg\min\limits_{\textbf{p}_m}\|{\textbf{p}_m}-{\textbf{p}_i}\|_2^2},\, i\neq m \rbrace, \\
		\widehat{\mathcal{Q}}=\lbrace{\widehat{\textbf{q}}}_j\mid{{\widehat{\textbf{q}}}_j=\arg\min\limits_{\textbf{q}_m}\|{\textbf{q}_m}-{\textbf{q}_j}\|_2^2},\, j\neq m \rbrace.
	\end{aligned}
\end{equation}

We construct a local patch $\mathcal{N}_{\circ}$ with $K$-nearest neighborhood for each point 
and extract local features with Dynamic Graph CNN \cite{wang2019dynamic} for input point cloud and 1-NN point cloud. Associated learned features are denoted as $\mathcal{F}_\mathcal{P}$, $\mathcal{F}_{\widehat{\mathcal{P}}}$, $\mathcal{F}_\mathcal{Q}$ and $\mathcal{F}_{\widehat{\mathcal{Q}}}$, respectively.
Then, matching scores are calculated in pairs with the normalized negative feature distance for input point cloud and 1-NN point cloud:
\begin{equation}
	\begin{aligned}
		\mathbf{M}_{i,j}=\text{softmax}\left( -\mathbf{D}_{i,1}, -\mathbf{D}_{i,2}, \dots, -\mathbf{D}_{i,M}\right) _{j},\\	
		\widehat{\mathbf{M}}_{i,j}=\text{softmax}\left( -\widehat{\mathbf{D}}_{i,1}, -\widehat{\mathbf{D}}_{i,2}, \dots, -\widehat{\mathbf{D}}_{i,M}\right) _{j},	
	\end{aligned}
	\label{nnff}
\end{equation}
where $\mathbf{D}_{i,j}=\|{\mathcal{F}_{\textbf{p}_i}-\mathcal{F}_{\textbf{q}_j}}\|_2$
and $\widehat{\mathbf{D}}_{i,j}=\|{\mathcal{F}_{\widehat{\textbf{p}}_i}-\mathcal{F}_{\widehat{\textbf{q}}_j}}\|_2$ denote the Euclidean distance between  learned local features. 
Based on the matching scores, we construct dual neighborhood matching map by fusing and averaging matching scores of each $\mathcal{N}_{\circ}$:
\begin{equation}
	\mathbf{G}_{i,j}=\frac{1}{K}\sum_{\textbf{p}_{i^{\prime}}\in\mathcal{N}_{\textbf{p}_i}}\sum_{\textbf{q}_{j^{\prime}}\in\mathcal{N}_{\textbf{q}_j}}\mathbf{M}_{{i^{\prime}},{j^{\prime}}} + \frac{1}{K}\sum_{\widehat{\textbf{p}}_{i^{\prime}}\in\mathcal{N}_{\widehat{\textbf{p}}_i}}\sum_{\widehat{\textbf{q}}_{j^{\prime}}\in\mathcal{N}_{\widehat{\textbf{q}}_j}}\widehat{\mathbf{M}}_{{i^{\prime}},{j^{\prime}}}.
	\label{finalmatching}
\end{equation}
%The higher matching score means consistently tend to have larger matching probabilities and the proposed module can provide higher quality matching than single neighborhood.
A higher matching score indicates consistently larger matching probability, thus highlighting the superior matching quality offered by 1-NN compared to single neighborhood.
The explanation is as follows: 
at least one matching is a correct matching in $\mathbf{M}_{i,j}$ and $\widehat{\mathbf{M}}_{i,j}$, thus leading to a high dual neighborhood matching score in $\mathbf{G}_{i,j}$. Even in cases where $\mathbf{M}_{i,j}$ and $\widehat{\mathbf{M}}_{i,j}$ are all false matching, the dual neighborhood matching score reflected in $\mathbf{G}_{i,j}$ is typically quite small and the real impact is limited.  
Finally, we formulate the final matching map as:
\begin{equation}
	\begin{aligned}
		\mathbf{F}_{i,j}=&\text{softmax}\left( -\mathbf{D}^{\prime}_{i,1}, -\mathbf{D}^{\prime}_{i,2}, \dots, -\mathbf{D}^{\prime}_{i,M}\right) _{j},\\	
		\mathbf{D}^{\prime}_{i,j}=&\exp{\left( \alpha-\mathbf{G}_{i,j}\right) *\left( \mathbf{D}_{i,j}+\widehat{\mathbf{D}}_{i,j}\right) },
	\end{aligned}
\end{equation}
where $\mathbf{D}^{\prime}_{i,j}$ is negatively related to the matching score  $\mathbf{G}_{i,j}$. We utilize the exponential strategy to control the changing ratio, along with a hyper-parameter $\alpha$ to control the influence of the dual neighborhood matching \cite{shen2022reliable}. Based on the high quality matching map $\mathbf{F}\in\mathbb{R}^{N\times M}$, we generate a reference point cloud copy
$\widetilde{\mathcal{Q}}\in\mathbb{R}^{N\times 3}$ including matching for each point ${\textbf{p}}_i$ in source point cloud $\mathcal{P}$:
\begin{equation}
	\widetilde{\textbf{q}}_i=\sum_{j=1}^{M}\mathbf{F}_{i,j} \cdot {\textbf{q}}_j.
	\label{generatepseudo}
\end{equation}
Note that Equation \ref{generatepseudo} means that reference point cloud copy $\widetilde{\mathcal{Q}}$ contains all predictive correspondences of 
${\textbf{p}}_i$ in source point cloud. Benefiting from the high quality matching map $\mathbf{F}\in\mathbb{R}^{N\times M}$, reference point cloud copy is also high quality and provides excellent precondition for inlier estimation. 
In particular, if ${\textbf{p}}_i$ is an inlier, the estimated correspondence $\widetilde{\textbf{q}}_i$ will appear in the correct position in reference copy, providing convenience for estimating inlier by capturing geometric structure neighborhood consistency. Conversely, if ${\textbf{p}}_i$ is an outlier, the estimated correspondence tends to have an unstable position.

\subsection{Geometric Neighborhood Inlier Estimation Module}
\label{gnie}
\begin{figure}
	\centering
	\includegraphics[width=3.4in]{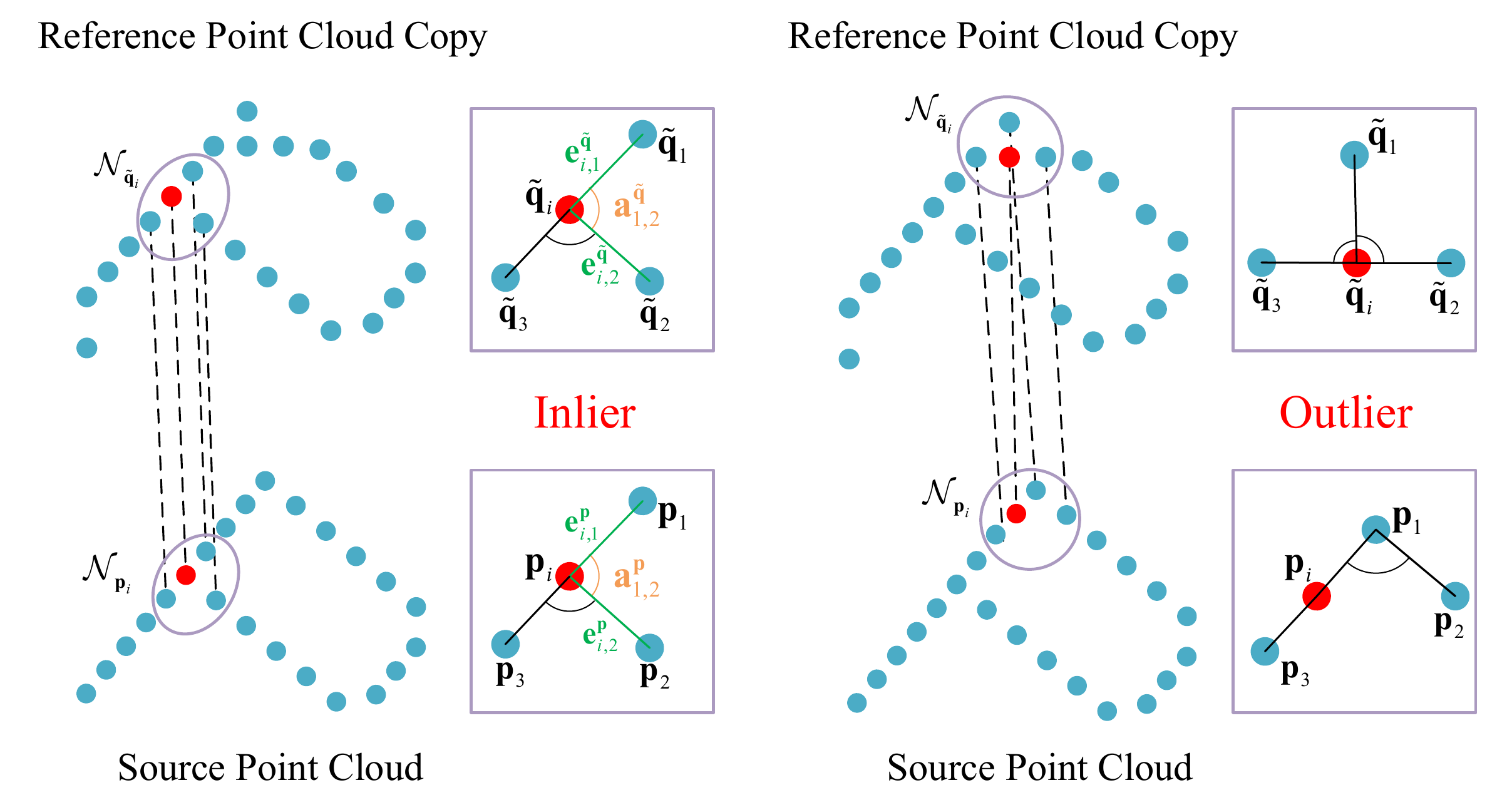}
	\caption{A toy example of inlier and outlier. The neighborhood graph formed by inlier and its neighborhood points should have geometric structure consistency (edge and included angle) between source point cloud and its reference point cloud copy. The outlier is just the opposite.} 
	\label{inliereval}
\end{figure}

In this section, we explore the geometric structure neighborhood consistency between the source point cloud and its reference point cloud copy for reliable inlier estimation.
As shown in Figure \ref{inliereval}, the neighborhood graph formed by inlier and its neighborhood points ($\mathcal{N}_{\textbf{p}_i}$ and $\mathcal{N}_{\widetilde{\textbf{q}}_i}$) should have consistency between source point cloud and its reference copy. Conversely, the neighborhood graph formed by outlier has significantly different geometric structure between $\mathcal{N}_{\textbf{p}_i}$ and $\mathcal{N}_{\widetilde{\textbf{q}}_i}$. Since estimated correspondence $\widetilde{\textbf{q}}_i$ tends to have an unstable position as illustrated in Section \ref{nfmm}, resulting in a chaotic neighborhood. Based on the above observation, we propose a geometric neighborhood inlier estimation module to construct effective transformation-invariant geometric structure representations, and adaptively capture the geometric structure consistency between $\mathcal{N}_{\textbf{p}_i}$ and $\mathcal{N}_{\widetilde{\textbf{q}}_i}$ to score the inlier confidence for each estimated correspondences. 

We first construct a learnable neighborhood graph by transformation-invariant geometric structure representations of $\mathcal{N}_{\textbf{p}_i}$ and $\mathcal{N}_{\widetilde{\textbf{q}}_i}$, which consists of edge representation and angle representation:
\begin{equation}
	\label{edgeangle}
	\begin{aligned}
		&\mathbf{e}_{i,k}^{\textbf{p}}=\textbf{p}_i-\textbf{p}_k, \; \mathbf{a}_{r,s}^{\textbf{p}}=\angle\left( \mathbf{e}_{i,r}^{\textbf{p}},\; \mathbf{e}_{i,s}^{\textbf{p}}\right) , \\	
		&\mathbf{e}_{i,k}^{\widetilde{\textbf{q}}}=\widetilde{\textbf{q}}_i-\widetilde{\textbf{q}}_k,	\;
		\mathbf{a}_{r,s}^{\widetilde{\textbf{q}}}=\angle\left( \mathbf{e}_{i,r}^{\widetilde{\textbf{q}}},\; \mathbf{e}_{i,s}^{\widetilde{\textbf{q}}}\right) ,\; \\ & i\neq k  ,\; k=1,\dots, K, r, s \in \left\{1,\dots, K\right\}
	\end{aligned}
\end{equation}
where $\textbf{p}_k$ and $\widetilde{\textbf{q}}_k$ are the points in $\mathcal{N}_{\textbf{p}_i}$ $\mathcal{N}_{\widetilde{\textbf{q}}_i}$. The numerically robust operator $\angle\left( \cdot_1,\cdot_2\right) $ computerd as:
\begin{equation}
	\angle\left(\cdot_1,\cdot_2\right)=\text{atan2}\left( \|{\cdot_1\times\cdot_2}\|,(\cdot_1)\cdot
	(\cdot_2)\right) ,
\end{equation}
which provides results in range $[0,\pi)$.
Moreover, these representations express sensitive and discriminative geometric structure in the point cloud, and provide adequate geometric cues for subsequent pipeline.

In order to better capture the neighborhood relevance and promote contextual message propagation, we utilize Multilayer Perceptron (MLP) $f_\theta$ with parameters $\theta$ to fuse representations and characterize the consistency between the neighborhoods
by the subtraction of the fused geometric representations:
\begin{equation}
	\mathbf{d}_{i,k}=f_\theta\left(\text{concat}\left(\mathbf{e}_{i,k}^{\textbf{p}}, \mathbf{a}_{r,s}^{\textbf{p}} \right) \right) - f_\theta\left(\text{concat}\left(\mathbf{e}_{i,k}^{\widetilde{\textbf{q}}}, \mathbf{a}_{r,s}^{\widetilde{\textbf{q}}} \right) \right). 
\end{equation}
Next, we further adaptively learn the attention coefficients $\delta_{i,k}$ of each geometric structure consistency:
\begin{equation}
	\delta_{i,k}=\text{softmax}\left(f_\mu\left(\mathbf{d}_{i,1}\right), f_\mu\left(\mathbf{d}_{i,2}\right), \dots, f_\mu\left(\mathbf{d}_{i,K}\right)  \right)_k,
\end{equation}
where $f_\mu$ is another MLP with parameters $\mu$. Then, we calculate the inlier confidence $\mathbf{w}_i$ of correspondence $({\textbf{p}}_i,\widetilde{\textbf{q}}_i)$ by aggregating the geometric structure consistency weighted:
\begin{equation}
	\mathbf{w}_i=1-\text{Tanh}\left(\left| l\left( \sum_{k=1}^{K}\delta_{i,k}\; *\; \mathbf{d}_{i,k} \right) \right|   \right), 
\end{equation}
where \textit{l} is a linear function. 
Finally, we select the largest $\mathit{N}_c$ weights as reliable inliers correspondence of the source and its reference copy:
\begin{equation}
	\mathcal{C}_h=\left\lbrace({\textbf{p}}_h,\widetilde{\textbf{q}}_h)\mid h\in \text{topk}\left(\mathbf{w}_i \right), i=1,\dots,N_c  \right\rbrace.
	\label{rics}
\end{equation}
We can solve transformation
$\lbrace\textbf{R}_{est},\textbf{t}_{est}\rbrace$ in closed form using weighted SVD based on reliable inliers correspondence, which has been shown to be differentiable in \cite{papadopoulo2000estimating}:
\begin{equation}
	\textbf{R}_{est},\textbf{t}_{est}=\min_{\textbf{R},\textbf{t}}\sum_{(\textbf{p}_i,\widetilde{\textbf{q}}_i)\in
		\mathcal{C}_h}\mathbf{w}_i\|{\textbf{R}\cdot\textbf{p}_i+
		\textbf{t}-\widetilde{\textbf{q}}_i}\|_2^2.
\end{equation}
We utilize an iterative scheme to update the source point cloud $\mathcal{P}$ with $\lbrace\textbf{R}_{est},\textbf{t}_{est}\rbrace$.

Besides, this inlier estimation method provides useful byproduct for unsupervised learning, which can be reliable self-supervised signal. More details can be seen in the following section.

\subsection{Optimization}
We conduct loss function for model optimization, which consists of four part. Then, we train the proposed model in an unsupervised manner instead of using the ground-truth transformations.

\textbf{Global Consistency Loss.} We investigate the global consistency loss between the final transformed source point cloud $\mathcal{P}^{\prime}$ and the reference point cloud $\mathcal{Q}$. 
We utilize the Huber function to assemble the global consistency loss, which is defined as follow:
\begin{equation}
	\mathcal{L}_{gc}=\sum_{\textbf{p}^{\prime}\in{\mathcal{P}^{\prime}}}H_{\beta}\left( \min_{\textbf{q}\in\mathcal{Q}}
	\|\textbf{p}^{\prime}-\textbf{q}\|_2^2\right) 
	+\sum_{\textbf{q}\in{\mathcal{Q}}}H_{\beta}\left( \min_{\textbf{p}^{\prime}\in\mathcal{P}^{\prime}}
	\|\textbf{q}-\textbf{p}^{\prime}\|_2^2\right).
\end{equation}
However, relying solely on global consistency loss is detrimental to the accuracy and reliability of our model.
Since the model may still potentially converge to sub-optimization due to the existing outliers, and massive potential information of point cloud is wasted. Hence, it is critical to mine the potential self-supervised signals in the point cloud and construct loss functions based on other existing elements.

\textbf{Dual Neighborhood Consistency Loss.} Based on reliable inliers correspondence in Equation \ref{rics}, we denote the inliers set of the source and its reference point cloud copy as $\textbf{X}\in\mathbb{R}^{\mathit{N}_c\times3}$ and $\textbf{Y}\in\mathbb{R}^{\mathit{N}_c\times3}$, respectively. We utilize
the neighborhood between the inliers to construct consistency objective, which aims to minimize the registration error between each neighborhood $\mathcal{N}_{\textbf{x}_i}$ and $\mathcal{N}_{\textbf{y}_i}$:
\begin{equation}
	\mathcal{L}_{in}=\sum_{\textbf{x}_i\in\textbf{X},\textbf{y}_i\in\textbf{Y}}
	\sum_{\textbf{p}_j\in\mathcal{N}_{\textbf{x}_i},\widetilde{\textbf{q}}_j\in\mathcal{N}_{\textbf{y}_i}}
	\left\| \textbf{R}_{est}\textbf{p}_j+\textbf{t}_{est}-\widetilde{\textbf{q}}_j\right\|_2,
\end{equation}
where $\mathcal{N}_{\textbf{x}_i}$ is transformed by $\mathcal{N}_{\textbf{y}_i}$. 

\textbf{Geometric Structure Consistency Loss.}
The geometric signal buried in point cloud is readily ignored, which hinders unsupervised inlier estimation. To address this issue, we design a geometric neighborhood loss with the reliable geometric self-supervised signal proposed in Section \ref{gnie}:
\begin{equation}
	\begin{aligned}
		\mathcal{L}_{gs}=&\sum_{\textbf{x}_i\in\textbf{X},\textbf{y}_i\in\textbf{Y}}
		\sum_{\textbf{p}_j\in\mathcal{N}_{\textbf{x}_i},\widetilde{\textbf{q}}_j\in\mathcal{N}_{\textbf{y}_i}}
		\left\| 
		\mathbf{e}_{i,j}^{\textbf{p}}-\mathbf{e}_{i,j}^{\widetilde{\textbf{q}}}
		\right\|_2+\\
		&\sum_{\textbf{x}_i\in\textbf{X},\textbf{y}_i\in\textbf{Y}}
		\sum_{\textbf{p}_j\in\mathcal{N}_{\textbf{x}_i},\widetilde{\textbf{q}}_j\in\mathcal{N}_{\textbf{y}_i}}
		\left\| 
		\mathbf{a}_{r,s}^{\textbf{p}}-\mathbf{a}_{r,s}^{\widetilde{\textbf{q}}}
		\right\|_2,
	\end{aligned}
\end{equation}
where $\mathbf{a}_{r,s}^{\textbf{p}}$ and $\mathbf{a}_{r,s}^{\widetilde{\textbf{q}}}$ is calculated by $\mathbf{e}_{i,j}^{\textbf{p}}$ and $\mathbf{e}_{i,j}^{\widetilde{\textbf{q}}}$ refering Equation \ref{edgeangle}.

\textbf{Spatial Consistency Loss.} We further explore to eliminate the spatial difference between the estimated correspondence and the real correspondence for each selected inlier $\textbf{x}_i$ and utilize spatial consistency loss with cross-entropy to sharpen matching map:
\begin{equation}
	\mathcal{L}_{sc}=-\frac{1}{\left| \textbf{X}\right|}\sum_{\textbf{x}_i\in\textbf{X}}\sum_{j=1}^{M}
	\llbracket j=\arg\max\limits_{j^{\prime}} \mathbf{G}_{i,j^{\prime}}\rrbracket \log\mathbf{G}_{i,j},
\end{equation}
where $\llbracket \cdot\rrbracket$ is the Iverson bracket. Spatial consistency loss
encourages to improve the matching probability and thus the estimated correspondence point in reference copy tends to have an stable position.

Since our work utilize an iterative scheme, we compute the loss at each iteration $N_l$ and have the weighted sum loss:
\begin{equation}
	\mathcal{L}=\sum_{l=1}^{N_l}\left( \mathcal{L}_{gc}^l+\gamma\mathcal{L}_{in}^l+\rho\mathcal{L}_{gs}^l+\lambda\mathcal{L}_{sc}^l\right) ,
	\label{totalloss}
\end{equation}
where $\gamma$, $\rho$ and $\lambda$ are trade-off parameters to control  corresponding loss function.

%------------------------------------------------------------------------
\section{Experiments}

\subsection{Experimental Setup}
\label{experimental setup}
We evaluate the proposed method on synthetic datasets ModelNet40 \cite{wu20153d} and Augmented ICL-NUIM \cite{choi2015robust}, and real-world dataset 7Scenes \cite{shotton2013scene}. ModelNet40 contains 12,308 CAD models of 40 different object categories. Augmented ICL-NUIM consists of 1,478 synthetic model generated by applying data augmentation on original 739 scan pairs.
7Scenes is a generally used dataset of indoor environment with 7 scenes including Chess, Fires, Heads, Office, Pumpkin, RedKitchen and Stairs.  

We compare our method to traditional methods and recent learning-based methods. The traditional methods include ICP \cite{besl1992method}
, FGR \cite{huang2020feature} and FPFH + RANSAC \cite{fischler1981random}. The recent learning based methods include IDAM \cite{li2020iterative}, FMR
\cite{huang2020feature}, RPMNet \cite{yew2020rpm}, CEMNet \cite{jiang2021sampling} and RIE \cite{shen2022reliable}. 
For consistency
with previous work, we measure Mean Isotropic Error
(MIE) and Mean Absolute Error (MAE). 
All metrics should be zero if reference point cloud align to source point cloud perfectly.
\begin{figure*}
	\centering
	\includegraphics[width=7in]{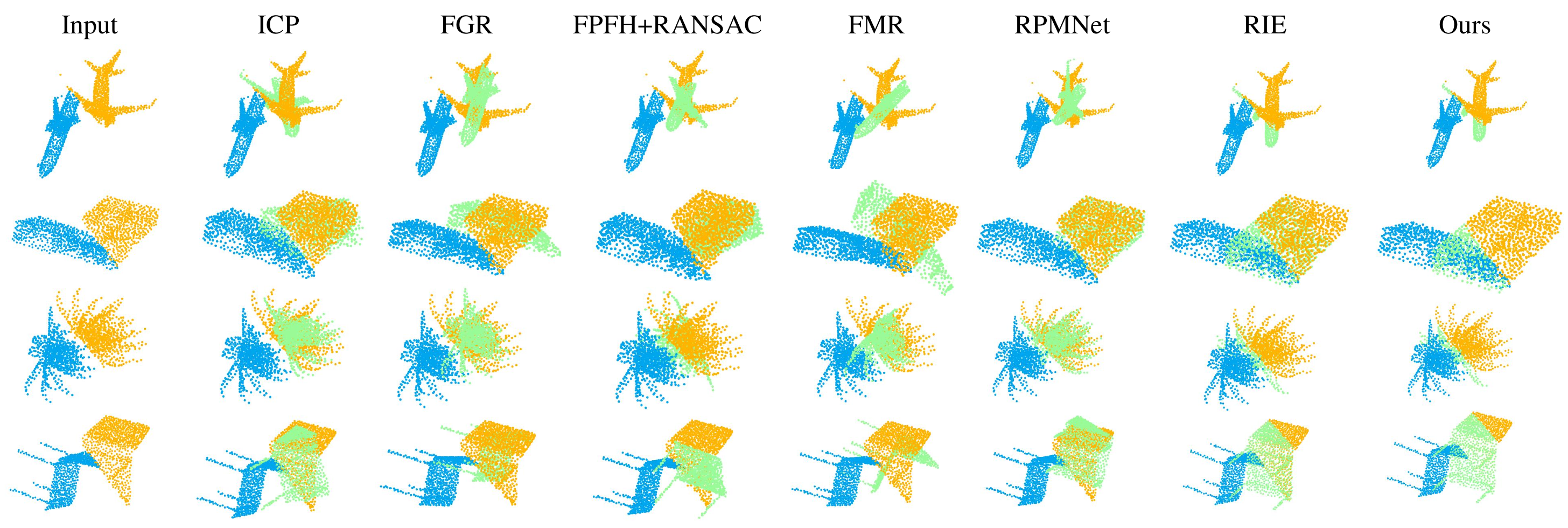}
	\caption{Qualitative comparison of 
		the registration results on unseen objects data
		(blue: source point cloud, yellow: reference point 
		cloud, green: transformed source point cloud). } 
	\label{visualmodelnet}
\end{figure*}

\begin{table*}
\begin{center}
	\footnotesize
	\renewcommand\tabcolsep{3.9pt}
	\begin{tabular}{l|cccc|cccc|cccc}
		\toprule 
		\multirow{2}{*}{Method} & \multicolumn{4}{c|}{Unseen Objects} 
		&\multicolumn{4}{c|}{Unseen Categories} & \multicolumn{4}{c}{Gaussian Noise}\\  
		&\makecell{MAE($\textbf{R}$)} & \makecell{MAE($\textbf{t}$)} & 
		\makecell{MIE($\textbf{R}$)}&\makecell{MIE($\textbf{t}$)}& 
		\makecell{MAE($\textbf{R}$)} & \makecell{MAE($\textbf{t}$)} & 
		\makecell{MIE($\textbf{R}$)}&\makecell{MIE($\textbf{t}$)}&
		\makecell{MAE($\textbf{R}$)} & \makecell{MAE($\textbf{t}$)} & 
		\makecell{MIE($\textbf{R}$)}&\makecell{MIE($\textbf{t}$)}\\
		\midrule
		%		\multirow{9}{*}{(A)} 
		ICP \cite{besl1992method} ($\bigcirc$)&3.4339&0.0114& 6.7706& 0.0227
		&3.6099 &0.0116& 7.0556& 0.0228&4.6441& 0.0167& 9.2194& 0.0333\\
		FGR \cite{huang2020feature} ($\bigcirc$)&0.5972& 0.0021& 1.1563 &0.0041
		&0.4579 &0.0016 &0.8442& 0.0032&1.0676& 0.0036& 2.0038 &0.0072\\
		FPFH+RANSAC \cite{fischler1981random} ($\bigcirc$) &0.7031& 0.0025& 1.2772 &0.0050& 0.4427& 0.0021& 0.9447 &0.0043& 1.4316& 0.0061& 2.5345 &0.0120\\
		IDAM \cite{li2020iterative} ($\blacktriangle$)&0.4243& 0.0020& 0.8170& 0.0040& 0.4809& 0.0028& 0.9157 &0.0055& 2.3076& 0.0124& 4.5332 &0.0246\\
		RPMNet \cite{yew2020rpm}
		 ($\blacktriangle$)&0.0051 &\textbf{0.0000}& \underline{0.0201}& \textbf{0.0000}& 0.0064& 0.0001& \underline{0.0207}& \underline{0.0001}& 0.0075& \textbf{0.0000} &\underline{0.0221}& \underline{0.0001}\\
		FMR \cite{huang2020feature} ($\blacktriangle$) &3.6497& 0.0101& 7.2810& 0.0200 &3.8594& 0.0114& 7.6450& 0.0225 &18.0355 &0.0536 &35.7986 &0.1063\\
		CEMNet\cite{jiang2021sampling} ($\triangle$) &0.1385& \underline{0.0001} &0.2489& \underline{0.0002}& 0.0804& \underline{0.0002}& 0.1405 &0.0003 &10.7026& 0.0393& 21.1836 &0.0781\\
		RIE \cite{shen2022reliable} ($\triangle$)& \underline{0.0033}& \textbf{0.0000} &0.0210 &\textbf{0.0000}& \underline{0.0059} & \textbf{0.0000}& 0.0228& \underline{0.0001}& \underline{0.0069} &\underline{0.0001} &0.0230 &\underline{0.0001}\\
		\midrule
		Ours ($\triangle$) &\textbf{0.0006} &\textbf{0.0000} &\textbf{0.0195}& \textbf{0.0000} &\textbf{0.0007} &\textbf{0.0000}& \textbf{0.0182}& \textbf{0.0000}& \textbf{0.0006}& \textbf{0.0000}& \textbf{0.0193}& \textbf{0.0000}\\
		\bottomrule
	\end{tabular}
\end{center}
\caption{Evaluation results on ModelNet40. Bold indicates the best 
	performance and underline indicates the second-best performance. ($\bigcirc$), ($\blacktriangle$) and ($\triangle$) denote the traditional, supervised and unsupervised methods, respectively.}
	\label{result-modelnet40}
\end{table*}

\subsection{ModelNet40 Dataset and Evaluation}
We first evaluate registration on ModelNet40, each point cloud 
contains 2,048 points that randomly sampled from  mesh faces and
normalized into a unit sphere. We randomly generate three Euler angle rotations
within $[0^\circ, 45^\circ]$ and translations within $[-0.5, 0.5]$ on each axis
as the rigid transformation during training.  Noted that, 
to simulate partial-to-partial registration, we crop the reference point cloud $\mathcal{Q}$ and the source point
cloud $\mathcal{P}$ respectively, and retain 70\% of the points.
All experiments on ModelNet40 take same settings.

\noindent\textbf{Unseen Objects.}
Our models are trained and tested on datasets comprising of samples belonging to the same categories, and both the training and test sets are obtained without any preprocessing or manipulation. We apply a random transformation on the reference point cloud 
$\mathcal{Q}$ to generate corresponding source point cloud $\mathcal{P}$.
Table \ref{result-modelnet40} shows quantitative results of the various algorithms 
under current experimental settings. The proposed method substantially outperforms all baseline in all metrics.  We can observe our method can even outperform the supervised IDAM, RPMNet and FMR by a large margin. 
Benefiting from the high quality reference point cloud copy and reliable inlier estimation, our method attains highly accurate registration and improves the registration accuracy by an order of magnitude.
In order to show the effect of our proposed approach clearly, a qualitative comparison of the registration results can be found in Figure \ref{visualmodelnet}. Our method ensure minimal impact on changing of the shape, and achieve the best performance even on asymmetric shape.

\noindent\textbf{Unseen Categories.}
To verify the generalization ability on categories, we train the models on the first 20 categories and test on the remaining unseen categories. 
The results are summarized in Table \ref{result-modelnet40}. We can observe that the majority of baseline consistently exhibit lower performance on the unseen categories, especially learning-based 
methods. In contrast, traditional algorithms are less susceptible to this issue due to the insensitivity of handcrafted methods to shape variance \cite{xu2021finet}.
Our registration process remains highly precise, achieving the lowest error across all metrics, while also maintaining acceptable levels of fluctuation.

\begin{figure}[h]
	\centering
	\includegraphics[width=3.3in]{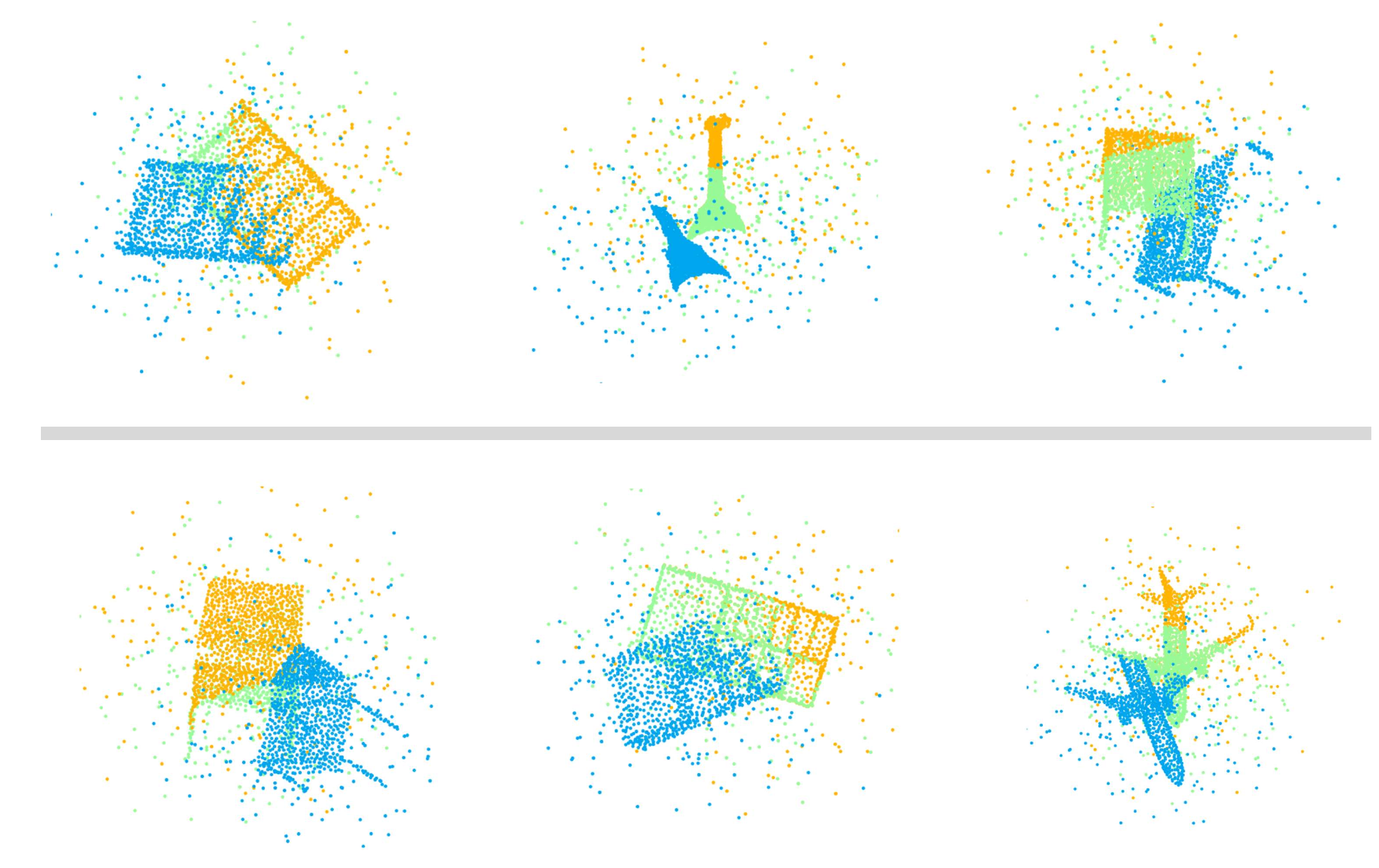}
	\caption{Visualization of registration results
		on the noisy ModelNet40 (blue: source point cloud, yellow: reference point 
		cloud, green: transformed source point cloud). The gray line is intended to separate the point cloud and prevent scattered points from impacting the clarity of visualization.}
	\label{vis-noise}
\end{figure}

\noindent\textbf{Gaussian Noise.}
In order to assess performance in the presence of noise, which is commonly encountered in real-world point clouds, we train our model on noise-free data and then evaluate all baseline using a test set featuring Gaussian noise. 
We randomly and independently generate noisy points to introduce noise into in source point cloud and reference point cloud by sampling from
$\mathcal{N}(0, 0.5)$ and clipped to $[-1.0, 1.0]$. 
This experiment is significantly more challenging, as constructing matching map and reference copy become much more difficult. As shown in Table \ref{result-modelnet40}, our method   
outperforms other baseline. In addition, a visualzation  
of registration results can be found in Figure \ref{vis-noise}. We can observe the Gaussian noise does not affect the registration of main body in point cloud. The experimental results is robust to the noise
and indirectly confirms the positive guiding effect of the 1-NN strategy on inlier estimation.

\begin{table*}
	\begin{center}
		\small
		\setlength{\tabcolsep}{5.3pt}
		\begin{tabular}{cccccc|cccc|cccc} 
			\toprule[1.5pt]
			\multirow{2}{*}{\makecell{DNFM \\Module}} & \multirow{2}{*}{\makecell{GNIE \\Module}} & \multirow{2}{*}{$\mathcal{L}_{gc}$} & \multirow{2}{*}{$\mathcal{L}_{in}$} & \multirow{2}{*}{$\mathcal{L}_{gs}$} & \multirow{2}{*}{$\mathcal{L}_{sc}$}& \multicolumn{4}{c|}{ModelNet40} & \multicolumn{4}{c}{7Sences}  \\
			&&&&&& MAE($\textbf{R}$)&MAE($\textbf{t}$) & MIE($\textbf{R}$)    &MIE($\textbf{R}$)& MAE($\textbf{R}$)&MAE($\textbf{t}$) & MIE($\textbf{R}$)    &MIE($\textbf{R}$) \\ 
			\midrule
			\CheckmarkBold& &\CheckmarkBold &\CheckmarkBold  &   &  \CheckmarkBold & 0.0919&0.0013 &0.1733&0.0026&3.4771&0.0195&7.0834&0.0368 \\
			&\CheckmarkBold &\CheckmarkBold & \CheckmarkBold & \CheckmarkBold  & \CheckmarkBold  &0.0106&0.0001&0.0271&0.0003&0.0051&\textbf{0.0000}&0.0240&\textbf{0.0001} \\
			\CheckmarkBold&\CheckmarkBold & &\CheckmarkBold  &  \CheckmarkBold & \CheckmarkBold  &0.0020& \textbf{0.0000} &0.0210& \textbf{0.0000}&0.0189&0.0002&0.0282&0.0003 \\
			\CheckmarkBold& \CheckmarkBold& \CheckmarkBold&  &  \CheckmarkBold & \CheckmarkBold  & 0.1774 &0.0018 & 0.2799 &0.0037 &0.8186&0.0194& 1.5426&0.0375 \\
			\CheckmarkBold& \CheckmarkBold& \CheckmarkBold& \CheckmarkBold  &   &\CheckmarkBold  & 0.0140 & 0.0002 &0.0343  & 0.0004&0.0111&0.0001& 0.0285&0.0002 \\
			\CheckmarkBold& \CheckmarkBold& \CheckmarkBold& \CheckmarkBold  &  \CheckmarkBold &  &0.0103 &0.0001&0.0281 & 0.0002&0.0262&0.0002& 0.0455&0.0004 \\
			\CheckmarkBold&\CheckmarkBold&\CheckmarkBold&\CheckmarkBold&\CheckmarkBold  &\CheckmarkBold&   \textbf{0.0006} &\textbf{0.0000} &\textbf{0.0195}& \textbf{0.0000} &\textbf{0.0036}& \textbf{0.0000}& \textbf{0.0191}& \textbf{0.0001} \\
			\bottomrule[1.5pt]
		\end{tabular}
	\end{center}
	\caption{Ablation study of different components on ModelNet40
		and 7Sences. DNFM Module: Dual Neighborhood Fusion Matching Module; GNIE Module: Geometric Neighborhood Inlier Estimation Module; {$\mathcal{L}_{gc}$}, {$\mathcal{L}_{in}$}, {$\mathcal{L}_{gs}$} and {$\mathcal{L}_{sc}$}: Each Loss Function in Equation \ref{totalloss}.}
	\label{ablation}
\end{table*}

\begin{table}
	\begin{center}
		\small
		\renewcommand\tabcolsep{3.4pt}
		\begin{tabular}{l|cccc}
			\toprule[1.5pt]
			%    \multicolumn{2}{c}{Part}                   \\
			%    \cmidrule(r){1-2}
			Method  & 
			MAE($\textbf{R}$) & MAE($\textbf{t}$) & 
			MIE($\textbf{R}$) & MIE($\textbf{t}$)\\
			\midrule
			\multicolumn{5}{c}{Augmented ICL-NUIM}\\
			\midrule
			ICP \cite{besl1992method} ($\bigcirc$)&2.4022& 0.0699 &4.4832 &0.1410\\
			FGR \cite{huang2020feature} ($\bigcirc$)&2.2477 &0.0808 &4.1850& 0.1573\\
			FPFH+RANSAC \cite{fischler1981random} ($\bigcirc$) &1.2349& 0.0429& 2.3167& 0.0839\\
			IDAM \cite{li2020iterative} ($\blacktriangle$)&4.4153& 0.1385& 8.6178& 0.2756\\
			RPMNet
			\cite{yew2020rpm} ($\blacktriangle$)& 0.3267 &0.0125& 0.6277& 0.0246\\
			FMR \cite{huang2020feature} ($\blacktriangle$) &1.1085& 0.0398& 2.1323& 0.0786\\
			CEMNet\cite{jiang2021sampling} ($\triangle$) &0.2374& \underline{0.0005}& 0.3987& \underline{0.0010} \\
			RIE \cite{shen2022reliable} ($\triangle$)& \underline{0.0492} &0.0023& \underline{0.0897}& 0.0049\\
			\midrule
			Ours ($\triangle$)&\textbf{0.0005}& \textbf{0.0002} &\textbf{0.0210} &\textbf{0.0004}\\
			\midrule
			\multicolumn{5}{c}{7Scenes}\\
			\midrule
			%		\multirow{9}{*}{(B)} 
			ICP \cite{besl1992method} ($\bigcirc$)& 6.0091& 0.0130 &13.0484& 0.0260\\
			FGR \cite{huang2020feature} ($\bigcirc$)&0.0919& 0.0004& 0.1705& 0.0008\\
			FPFH+RANSAC \cite{fischler1981random} ($\bigcirc$) &1.2325& 0.0062& 2.1875& 0.0124\\
			IDAM \cite{li2020iterative} ($\blacktriangle$)&5.6727& 0.0303& 11.5949&0.0629\\
			RPMNet
			\cite{yew2020rpm} ($\blacktriangle$)& 0.3885& 0.0021& 0.7649 &0.0042\\
			FMR \cite{huang2020feature} ($\blacktriangle$) &2.5438 &0.0072& 4.9089& 0.0150\\
			CEMNet\cite{jiang2021sampling} ($\triangle$) &0.0559&  \underline{0.0001}& 0.0772&  \underline{0.0003}\\
			RIE \cite{shen2022reliable} ($\triangle$)&  \underline{0.0121} & \underline{0.0001}&  \underline{0.0299} &\textbf{0.0001}\\
			\midrule
			Ours ($\triangle$)&\textbf{0.0036}& \textbf{0.0000}& \textbf{0.0191}& \textbf{0.0001}\\
			\bottomrule[1.5pt]
		\end{tabular}
	\end{center}
	\caption{Evaluation results on Augmented ICL-NUIM and 7Scenes. Bold indicates the best performance and underline indicates the second-best performance. ($\bigcirc$), ($\blacktriangle$) and ($\triangle$) denote the traditional, supervised and unsupervised methods, respectively.}
	\label{result-other}
\end{table}

\subsection{Other Datasets and Evaluation}
We further conduct comparison evaluation on other datasets: 7Scenes and Augmented ICL-NUIM. We sample the reference point clouds to 2,048 points and randomly sample three Euler angle rotations within $[0^\circ, 45^\circ]$ and translations within $[-0.5, 0.5]$ on each axis as the rigid transformation to obtain source point clouds, then downsample the point clouds to 1,536 points to generate the partial data. As demonstrated in Table \ref{result-other}, our method exhibits extremely higher registration precision on all criteria on Augmneted ICL-NUIM and 7Scenes, especially the rotation error. 
Due to space limit, we present more visualization 
results and quantitative comparison results in appendix. 
We can summarize our method has best performance and is comfortable with real-world dataset.

\subsection{Ablation Study and Analysis}
In this section, we conduct extensive ablation studies for a better understanding of the various modules in our method on ModelNet40 and 7Scenes. Due to space limit, we present more ablation studies in appendix.

\noindent \textbf{Dual Neighborhood Fusion Matching Module.} We first conduct ablation
study on proposed 1-NN point cloud in dual neighborhood fusion matching module. We replace this component with single neighborhood. As shown in the second and seventh rows of Table \ref{ablation}, applying solely the single neighborhood brings no performance gain, since the single neighborhood confuse the matching map and thereby lower the quality of the generated reference point cloud copy. Therefore, we fuse 1-NN point cloud to enhance neighborhood matching map which can promote the generation of correct reference point cloud copy.

\noindent \textbf{Geometric Neighborhood Inlier Estimation Module.} We further evaluate the effect of geometric neighborhood inlier estimation module. In this experiment, we do not construct transformation-invariant geometric structure representations for neighborhood but rather utilize raw coordinate to estimation inlier. As shown in the first and seventh rows of Table \ref{ablation}, we can observe coordinate-based method has significantly degraded performance, because coordinate-based method
cannot provide transformation-invariant representations, leading to high stochasticity in inlier estimation.
Noted that, geometric neighborhood inlier estimation module provides a self-supervised signal for our model, and $\mathcal{L}_{gs}$ is designed based on this signal. Therefore, when geometric neighborhood is deleted, the loss function $\mathcal{L}_{gs}$ should not participate in optimization.

\noindent \textbf{Loss Function.} Comparing 3$\sim$6 rows in Table \ref{ablation}, we evaluate
the performance of the model with different loss functions.
Comprehensively, lacking any part of $\mathcal{L}$ will degrade the
performance of the model. 
The error observed in the fourth row is extremely large, primarily due to the absence of any optimization objective related to transformation in the loss function. This significantly reduces the registration performance of the model.
In particular, this ablation study confirms our prediction of $\mathcal{L}_{gs}$ and $\mathcal{L}_{sc}$ can provide reliable and effective self-supervised signal and improve the matching map construction.

%------------------------------------------------------------------------
\vspace{-0.1cm}
\section{Conclusion}
We propose an effective inlier estimation method for unsupervised point cloud registration, which aims to capture geometric structure consistency between source point cloud and its corresponding reference point cloud copy. We design 1-NN point cloud to potentially facilitate matching map construction for obtaining high quality reference copy. Based on the high quality reference copy, and observation that the neighborhood graph formed by inlier and its neighborhood points should have geometric structure consistency between source and its reference copy, we design a geometric neighborhood inlier estimation module to score the inlier confidence for each estimated correspondence and provide simultaneously the effective self-supervised signal based on geometric structure for model optimization. Finally, we conduct extensive experiments on  ModelNet40, Augmented ICL-NUIM and 7Scenes, demonstrating that our unsupervised framework can achieve outstanding performance and 1-NN strategy effectively guides inlier estimation. Moreover, the visualizations of complete
predictions demonstrate that the results are faithful and plausible.

%------------------------------------------------------------------------
{\small
\bibliographystyle{ieee_fullname}
\bibliography{references}
}

\end{document}